\documentclass[lettersize,journal]{IEEEtran}
\usepackage{amsmath}
\usepackage{amsfonts}
\usepackage{amssymb}
\usepackage{graphicx}
\usepackage{multirow}
\usepackage[dvipsnames]{xcolor}
\usepackage[nocompress]{cite}
\usepackage[caption=false,font=normalsize,labelfont=sf,textfont=sf]{subfig}
\usepackage{pbalance}

\usepackage{comment}

\newcommand*{\nnew}[1]{\textcolor{black}{#1}}
\newcommand*{\nnnew}[1]{\textcolor{black}{#1}}

\begin{document}

\newcommand{\vc}[1]{\bar{\mathbf{#1}}}
\newcommand{\vchat}[1]{\hat{\mathbf{#1}}}
\newcommand{\vctilde}[1]{\tilde{\mathbf{#1}}}
\newcommand{\set}[1]{\mathbf{#1}}
\title{
Large-Margin Hyperdimensional Computing:\\A Learning-Theoretical Perspective
}

\author{%
Nikita~Zeulin, Olga~Galinina, Ravikumar~Balakrishnan, Nageen~Himayat, and~Sergey~Andreev%
\IEEEcompsocitemizethanks{\IEEEcompsocthanksitem N. Zeulin and S. Andreev are with  Tampere University, Tampere, Finland.\protect\\
E-mail: \{nikita.zeulin,sergey.andreev\}@tuni.fi
\IEEEcompsocthanksitem O. Galinina is with Tampere Institute for Advanced Study, Tampere University, Finland. E-mail: olga.galinina@tuni.fi.\protect\noindent
\IEEEcompsocthanksitem N. Himayat and R. Balakrishnan are with Intel Corporation, Santa Clara, CA, United States. E-mail: \{nageen.himayat,ravikumar.balakrishnan\}@intel.com}\protect\\
}

\markboth{}%
{Zeulin \MakeLowercase{\textit{et al.}}: Large-Margin Hyperdimensional Computing: A Learning-Theoretical Perspective}

\IEEEtitleabstractindextext{%
\begin{abstract}

Overparameterized machine learning (ML) methods such as neural networks may be prohibitively resource-intensive for devices with limited computational capabilities. Hyperdimensional computing (HDC) is an emerging resource-efficient and low-complexity ML method that allows hardware-efficient implementations of (re-)training and inference procedures. In this paper, we propose a maximum-margin HDC classifier, which significantly outperforms baseline HDC methods on several benchmark datasets. Our method leverages a formal relation between HDC and support vector machines (SVMs) that we established for the first time. Our findings may inspire novel HDC methods with potentially more hardware-oriented implementations compared to SVMs, thus enabling more efficient learning solutions for various intelligent resource-constrained applications.
\end{abstract}

}

\maketitle

\IEEEdisplaynontitleabstractindextext

\section{Introduction}
\subsection{Motivation}
\IEEEPARstart{O}{ver} the past years, integration of machine learning~(ML) algorithms into various downstream tasks has been an active topic in many research fields. Specifically, there has been a remarkable advancement in the development of methods based on large, overparameterized neural networks. However, such methods may involve prohibitively high computational complexity for real-time training or inference in applications with inherent resource constraints. 

Hyperdimensional computing (HDC)~\cite{kanerva2009hyperdimensional} provides a conceptually different perspective on pattern recognition. It relies on a distributed, or holographic, representation of the data, where information is distributed over thousands of dimensions~\cite{kleyko2024design}. HDC represents original low-dimensional data points as very large, randomized hypervectors, typically consisting of $5\text{-}10$K real-, integer-, or binary-valued features. Such distributed representations make the hypervectors robust to individual position errors\nnew{, which is exploited for improving fault-tolerance of classification algorithms \cite{fan2024efficient}.}

\nnew{These core principles of HDC 
have motivated a large body of work on hardware-efficient implementations compared to deep neural networks (DNNs).} While neural networks rely on resource-intensive matrix-to-matrix multiplications, the training operations in HDC can be reduced to element-wise summations, multiplications, or dot-products over the individual hypervectors. Moreover, HDC natively supports online retraining, which can be beneficial for applications with online learning on resource-constrained devices. HDC has been successfully adapted for \nnew{the classification of images \cite{dutta2022hdnn, pourmand2025laplace, smets2024encoding, moran2019energy}}, time series~\cite{schlegel2022hdc, moreno2024kalmanhd}, graphs \cite{nunes2022graphhd, kang2025relhdx}, federated learning~\cite{ergun2023federated, tian2025federated, zeulin2023resource}, medical applications \cite{kleyko2019deep}, and other problems. \nnew{A comprehensive overview of the existing HDC classification algorithms is provided in the recent survey \cite{verges2025classification}.}

A more traditional alternative to overparameterized deep learning methods is support vector machines (SVMs), a classical statistical learning framework. SVMs offer resource-efficient solutions and can successfully compete with DNNs across various tasks~\cite{liu2017svm}. Compared to DNNs, SVM models demonstrate better generalization ability even if trained on tens of data points and can offer training and inference that are orders of magnitude faster~\cite{hua2023lab,carter2022fast}. As an example, SVM-based algorithms have been applied for positioning in networking applications \cite{hua2023lab}, on-device malware detection for industrial Internet of Things networks \cite{carter2022fast}, and accurate probabilistic load forecasting for energy storage systems \cite{zhang2023long}. Although SVM models typically utilize kernel functions to handle non-linearity, they also support DNN-based feature extractors for higher performance on complex data~\cite{tang2013deep}.

\nnnew{In contrast to the analytically rigorous SVM framework, many HDC methods rely on heuristic update rules with limited analytical justification. While many HDC-centric works focus on empirically successful hardware-level implementations, the theoretical foundations of HDC as a learning framework are still developing. In the following, we provide an overview of the previous efforts in this direction, together with several notable works in HDC research.}

\subsection{Related work}
{Originally, HDC was first introduced by P. Kanerva \cite{kanerva2009hyperdimensional} as an umbrella term for different vector symbol architectures (VSAs) \cite{gayler2004vector} that employ randomized, high-dimensional data representations. Conventional symbolic HDC frameworks, such as multiply-accumulate-permute \cite{gayler1998multiplicative}, binary spatter codes \cite{kanerva1994spatter}, or Fourier holographic reduced representations (FHRR) \cite{plate1994distributed}, were adopted to formally encode relations between independent entities and enable analytically explainable reasoning via similarity checks between hypervectors. 
\nnew{Comprehensive overviews of VSA models, data transformations, and applications can be found in \cite{kleyko2023survey1, kleyko2023survey2}.}

{HDC has become more closely associated with hardware-efficient pattern recognition frameworks, while VSAs are more often discussed in the context of symbolic representations \cite{kleyko2018classification}. This distinction became more pronounced after HDC was adapted to the classification of texts \cite{rahimi2016robust} and electromyography signals \cite{rahimi2016hyperdimensional}. These early works demonstrated the simplicity of hardware implementation of HDC classifiers, while achieving performance comparable to or higher than established baselines.}

This discovery inspired a surge of interest in ML and hardware-oriented communities to develop both theoretical foundations and resource-efficient hardware implementations for  ML-related HDC frameworks. Below, we briefly review several representative efforts in HDC research related to HD representations, hardware-oriented implementations, \nnew{and studies that adapt HDC for other learning applications.}
\subsubsection{Structure-Preserving HD Representations}
{While conventional HDC frameworks can demonstrate high performance on textual and time-series data, they may be less suitable for structured data, such as images and graphs. In the case of visual pattern recognition, the conventional pixel-wise encodings become unsuitable as they do not provide translation invariance and treat individual pixels independently. \nnew{One of the first HDC frameworks to overcome this limitation was proposed in \cite{dutta2022hdnn}, where image hypervectors were constructed from features extracted with a convolutional neural network (CNN). With this approach, for the first time, HDC achieved comparable classification performance to conventional neural network-based baselines on such a complex image dataset as CIFAR-10 \cite{krizhevsky2009learning}.} However, this approach requires additional computational resources for running CNN inference and does not produce explainable feature representations.} 

{As an alternative, the work in \cite{pourmand2025laplace} proposes the Laplace-HDC framework that constructs translation-invariant hypervector representations based on the Laplace kernel. The method constructs a set of bipolar hypervectors from a semi-positive definite covariance matrix that meets a specific criterion. To support translation-invariance, this approach employs 2D cyclic permutations for binding with hypervectors representing the image (or one of its tiles).}

\nnew{Another line of work considers HD encoding methods for binary images. For example, the work in \cite{smets2024encoding} introduces the HD encoding that preserves similarity between hypervectors of spatially close points, while keeping more distant vectors orthogonal. The proposed method achieves state-of-the-art accuracy on MNIST and Fashion MNIST datasets; however, the input is currently limited to monochrome images.}

{Another approach for producing structure-aware hypervectors is to use kernel approximations and perform similarity checks between hypervectors in the kernel space. This approach was analytically studied in \cite[Sec. 5.2.1]{thomas2021theoretical}, where hypervectors were constructed using random Fourier features (RFF) that approximate the radial basis function kernel. 
We note, however, that a similar idea was concurrently introduced in \cite{zou2021scalable} and \cite{yu2022understanding} to capture non-linear relations between data features.} 

{While the RFF method is most commonly applied for approximating the shift-invariant radial basis function kernel, arbitrary kernels can be approximated with the Nystr\"{o}m method \cite{williams2000using}. The work in \cite{Zhao_Thomas_Brin_Yu_Rosing_2025} proposes the NysHD framework, which adapts the Nystr\"{o}m method to construct hypervectors approximating the kernel covariance matrix. This is a powerful method since it can approximate kernels for different types of data, such as strings, time series, and graphs. The Nystr\"{o}m method is, however, known to be sensitive to the selection of landmarks for the sub-kernel matrix and may be prone to numerical instability.}

Most existing studies focus on improving the quality of HD data representations. In contrast, our work analyzes the learning objective of HDC classification.

\subsubsection{\nnew{Hardware-Oriented HDC Research}}
\nnew{Although hardware-oriented HDC is not the focus of this work, we briefly summarize representative efforts for completeness.
}
{The work in \cite{liang2023dependablehd} introduces an in-memory HDC framework DependableHD, which is resilient to bit-level failures.
DependableHD supports a controlled number of misclassifications by introducing an adjustable classification margin and adding a small amount of noise to improve the classification performance. Adding noise during training is a well-known regularization mechanism that can improve generalization of the model \cite{bishop1995training}.} 
The DropHD framework~\cite{genssler2024drophd} 
exploits the fact that hypervector positions with low entropy are more prone to hardware errors and prunes them to improve the hypervector resilience. This approach can be applied for non-volatile memories with data-dependent, non-uniform level changes. 
{A similar idea was previously proposed in \cite{khaleghi2020prive}, where the resilience of HDC against model inversion attacks is enhanced by pruning the hypervector positions with minimal impact on the performance.}
The MicroHD framework~\cite{ponzina2024microhd} optimizes the model hyperparameters (hypervector size, bitwidth, and number of quantization levels for feature values) to minimize memory and compute consumption given the user-defined maximum accuracy degradation. \nnew{Furthermore, \cite{kleyko2022cellular} explores the use of cellular automata (CA) for on-the-fly generation of temporal features for reservoir computing and resource-efficient implementation of image classification \cite{moran2019energy}. }

\nnew{While these hardware-oriented studies focus on HDC implementation and robustness and are orthogonal to the proposed theoretical perspective, they illustrate the potential for the theory-based design of future accelerators.
}

\subsubsection{\nnew{HDC-Powered ML Frameworks}}
\nnew{There are several contributions that apply the HDC principles to learning tasks beyond conventional supervised learning.} For example, the reinforcement Q-learning-based approach in \cite{ni2024efficient} encodes states and actions into hypervectors and computes the Q-values using similarity checks between them. Action hypervectors can be subsequently refined based on discrepancies between true and predicted Q-values using the HDC retraining procedure. The work in \cite{angioli2025hd} incorporated HDC into a contextual multi-armed bandit setting with a linear relationship between the context and the reward. Similar to \cite{kang2025relhdx}, this framework encodes the context as a hypervector and performs similarity checks with the hypervectors of actions to infer the value of the payoff. Beyond that, the ideas of symbolic HDC representations have been adopted in computer vision frameworks to enhance out-of-distribution detection \cite{wilson2023hyperdimensional} and to improve uncertainty estimation in 3D object detection under visual impairments~\cite{chen2025hyperdimensional}. 

\nnew{These works
illustrate the use of HDC in ML tasks beyond standard classification. In contrast, our contribution targets the core supervised setting and strengthen its theoretical foundations by
relating HDC classifiers to linear soft-margin SVMs.}

\subsubsection{Relation Between HDC and Classical ML Methods}

\nnnew{One of the first efforts to establish theoretical foundations of HDC was presented in \cite{thomas2021theoretical}, with the focus on the analytical properties of hypervector representations. A formal perspective on the HDC training principles was established in the subsequent works by relating HDC to the existing learning frameworks. 
For example, the work in \cite{diao2021generalized} highlights the similarity between HDC and generalized learning vector quantization (GLVQ) \cite{sato1995generalized}. Particularly, this work demonstrates that the conventional perceptron-based HDC training procedure closely resembles the iterative gradient descent-based optimization of the GLVQ loss function.}

\nnnew{The work in \cite{ergun2023federated} establishes a relation between a binary HDC classifier and Fisher's linear discriminant. Similarly, the perceptron-based HDC training is shown to be identical to the gradient descent-based empirical risk minimization of the linear discriminant. In \cite{hernandez2024hyperdimensional}, the HDC-based regression method was derived from the kernel ridge regression, where the hypervectors were constructed using RFF~\cite{rahimi2007random} that approximate shift-invariant kernels. }

\nnnew{Our work continues this line of research by establishing a formal relationship between HDC classifiers and soft-margin SVMs. This perspective offers a theoretical basis for developing new HDC algorithms.}

\subsection{Our Contribution}

\nnnew{In this article, we establish a formal relation between binary (two-class) HDC and linear SVM classifiers. Our contributions are as follows:
\begin{itemize}
    \item We demonstrate that a binary HDC classifier can be reformulated as a special case of a linear SVM classifier with identical decision rules. 
    \item Leveraging the established relation, we propose an iterative algorithm to train a novel \textit{binary maximum-margin HDC (MM-HDC) classifier} with potentially higher generalization ability compared to conventional HDC\,methods. 
    \item  We present a detailed performance evaluation and demonstrate that our MM-HDC classifier outperforms the baseline HDC algorithms on the selected benchmark datasets.
\end{itemize}
}

\nnew{We emphasize that the discovered relation between HDC and SVM allows us to apply rich theoretical foundations of SVM to the formal analysis of HDC algorithms. Importantly, our MM-HDC framework reveals analytical justifications for several successful HDC methods, including OnlineHD \cite{hernandez2021onlinehd}, DependableHD \cite{liang2023dependablehd}, and LeHDC \cite{duan2022lehdc}. While HDC in general is often motivated by its lightweight and hardware-efficient implementations, this work is primarily theoretical and algorithmic. 
We expect that our findings can open new opportunities for designing new, analytically grounded HDC-based algorithms for well-established SVM tasks, such as regression and anomaly detection.}

\section{Background and Problem Formulation}

We consider a classification problem with a dataset $\{\set{X},\set{y} \}$ consisting of $N$ $d$-dimensional data points $\set{x}_j \in \set{X}$ and corresponding labels $y_j\in\{-1,+1\}$. This section briefly reviews HDC training and inference procedures and describes the SVM problem formulation. For convenience, we also summarize the employed notations in Table \ref{tab:notations}.

\begin{table}[t!]
    \centering
    \caption{Notations and descriptions used in this work}
    {
    \label{tab:notations}
    \begin{tabular}{@{}ll@{}}
    \hline
    \textbf{Notation} & \textbf{Description} \\ \hline
    $N$ & Number of data points \\
    $d$ & Dimensionality of data points \\
    $\set{X} \in \mathbb{R}^{d \times N}\!$ & Set of $d$-dimensional data points \\
    $\set{y}  \in \mathbb{R}^{N}$ & Vector of corresponding labels, $y_j \in \{-1,+1\}$ \\
    $\set{x}_j \in \mathbb{R}^{d}$ & Individual $i$-th data point \\
    $y_j$ & Label of individual $i$-th data point \\
    $D$ & Dimensionality of hypervector \\
    $\theta(\cdot)$ & HD representation function, $\mathbb{R}^d \rightarrow \mathbb{R}^D$ \\   
    $p_i$ & Prototype of class $i$, $i\in\{-1,+1\}$ \\
    $\mathcal{C}_i$ & Set of indices of data points corresponding to class $i$  \\
    $a_j$ & Weighting coefficients for prototype initialization \\
    $\text{sim}(\set{a},\set{b})$ & Similarity function between vectors $\set{a}$ and $\set{b}$ \\
    $\alpha$ & Learning rate for prototype retraining \\
    $\set{w}$ & Separating hyperplane \\
    $b$ & Bias term \\
    $\zeta_i$ & Slack variable corresponding to $i$-th data point \\
    $C$ & Regularization constant \\
    $\lambda_i$ & Lagrange multiplier corresponding to $i$-th data point $\set{x}_i$ \\
    $\eta_i$ & Lagrange multiplier corresponding to $i$-th slack variable $\zeta_i$ \\
    $\mathcal{A}_+,\ \mathcal{A}_-$ & Sets of points within the positive and negative class margins \\
    $\mathcal{F}(\cdot)$ & Objective function for HDC optimization \\ \hline
    \end{tabular}
    }
\end{table}

\subsection{Binary HDC Classification}
\label{sec:hdc_classification}

{Most HDC algorithms include four core procedures: (i) data projection, (ii) prototype initialization, (iii) similarity check, and (iv) iterative prototype retraining. Below, we describe these procedures for a binary HDC classifier.}

{\textbf{Data projection.}} In HDC, each data point $\set{x}_j\in\set{X}$ is mapped to $D$-dimensional hyperdimensional (HD) space using a randomized mapping function $\theta(\set{x}_j)$: $\mathbb{R}^d\to\mathbb{R}^D$. \nnew{Without loss of generality, we consider a real-valued mapping for convenience in relating to the linear $C$-SVM formulation. Our analytical results can be readily extended to integer- or binary-valued hypervectors, as the proposed analysis does not impose any strict limitation on the choice of transform~$\theta(\cdot)$.} Therefore, the mapping can be selected based on the encoding complexity or desired analytical properties, as in symbolic HDC frameworks.

{\textbf{Prototype initialization.}} Let $\mathcal{C}_i$ represent a set of data indices that correspond to the $i\text{-th}$ class, $i\in\{-1, +1\}$ for binary classification. The training data of each class are bundled into the respective \textit{class prototype} $\set{p}_i$ defined as
\begin{equation}\label{eq:1}
    \set{p}_i = \sum_{j\in\mathcal{C}_i}a_j\theta(\set{x}_j), \quad i \in\{-1,+1\},
\end{equation}
where $a_j$ are the weighting coefficients. A typical heuristic is to set $a_j = \frac{1}{|\mathcal{C}_i|}$, which means that the prototype is initialized as the average of HD representations $\theta(\set{x}_j)$ of the data points $\set{x}_j$ in the respective class. The exact values of the coefficients $\{a_j\}$ are implicitly refined during the HDC retraining procedure, as we demonstrate below. {The prototypes can be interpreted as an ``average'' HD representation of the class data. Depending on the representation type, the prototypes can be converted back into the original data space \cite{khaleghi2020prive}.}

{\textbf{Similarity check and inference.}} The HDC inference procedure involves computing similarity values between the HD representation $\theta(\set{x}_*)$ of the test point $\set{x}_*$ and the prototype~${\set{p}_i}$. The similarity between $\set{a}$ and $\set{b}$ is denoted as $\mathrm{sim}(\set{a},\set{b})$. {For real-valued vectors, a typical choice of similarity is the dot product or cosine similarity, depending on the application.} Therefore, for a point $\set{x}_*$, its label $\hat{y}_*$ may be estimated as
\begin{equation}\label{eq:hdc_class_rule}
   \!\! \hat{y}_* \!= \!\arg\max_i \mathrm{sim}(\theta(\set{x}_*),\set{p}_i).
\end{equation}

{\textbf{Iterative prototype retraining.}} To further improve the classification performance after the prototype initialization in~\eqref{eq:1}, one can perform a \textit{retraining procedure}. The retraining algorithm exploits the fact that some data points may have substantial similarity despite belonging to different classes, thus resulting in class prototypes that can be characterized as ``fuzzy''. Therefore, the goal of the retraining procedure is to increase dissimilarity between the prototypes by reinforcing correct predictions while penalizing incorrect ones. If a training data point $\set{x}$ of class $i$ is misclassified as class $j$, then the corresponding HDC prototypes $\set{p}_i$ and $\set{p}_j$ are updated as 
\begin{gather}\label{eq:upd_rule}
    \set{p}_i = \set{p}_i + \alpha \cdot \theta(\set{x}),~\set{p}_j = \set{p}_j - \alpha \cdot \theta(\set{x}),
\end{gather}
where $\alpha\in[0,1]$ is the learning rate. We refer to the prototype update rule in \eqref{eq:upd_rule} as the \textit{perceptron-based retraining procedure}, which was originally employed for HDC training and further developed in subsequent works~\cite{hernandez2021onlinehd}. 

\nnew{The specific implementation of the retraining procedure in \eqref{eq:upd_rule} largely depends on the employed hypervector type, which can be binary-, integer-, or real-valued. Particularly, the implementations based on floating-point arithmetics require minimal or no care of overflow/underflow or normalization. However, they are less spread in practice due to lower computational and energy efficiency compared to the fixed-point counterparts. Nevertheless, the performance of real-valued implementations is close to the integer-based ones and can be viewed as an upper bound for achievable classification performance~\cite{thomas2021theoretical}.}

\nnew{To further enhance the predictive performance of HDC classifiers, previous studies proposed several algorithmic enhancements to the discussed retraining procedure. For example, the OnlineHD algorithm \cite{hernandez2021onlinehd} implements the adaptive retraining algorithm based on \eqref{eq:upd_rule} by weighting the misclassified hypervectors with similarity values corresponding to the refined prototypes.} The retraining procedure of LeHDC \cite{duan2022lehdc} refines every prototype with higher similarity than that of the correct-class prototype. The DependableHD framework \cite{liang2023dependablehd} introduces the similarity margin that controls the maximum possible violation of the similarity check. \nnew{However, the modifications presented in these studies, albeit effective, are heuristic and lack formal theoretical justification. These enhancements can be analytically explained within the large-margin HDC classification framework presented in this work.}

\subsection{Binary Linear $C$-SVM Classification}

In the binary SVM classification, the objective is to find a hyperplane that separates the training points of two classes, while maximizing the distance from the nearest points of each class. We assume that the separating hyperplane is defined by equation $ {\langle\set{w}, \set{x}\rangle} = b$, $\set{w} \in \mathbb{R}^d$, $b \in \mathbb{R}$. Then, the label $y_i$ of a data point $\set{x}_i \in \set{X}$ can be estimated using the relative position w.r.t. the hyperplane, which is equivalent to the following decision rule:
\begin{equation}\label{eq:svm_decision_rule}
\hat y_i  = \operatorname{sign}\left(\langle\set{x}_i,\set{w}\rangle -b \right).
\end{equation}
The parameters $\set{w},b$ of the separating hyperplane can be arbitrarily scaled by a positive constant. Therefore, it is commonly assumed that 
for the nearest points, or \textit{support vectors}, ${y_i\cdot \left[\langle\set{x}_i,\set{w}\rangle -b \right] = 1}$.

In the case of a linearly separable training set, the following conditions should hold for every point $\set{x}_i$:
\begin{equation} \label{eq:svm_support_cond}
\left\{ 
    \begin{array}{ll}
    \langle\set{x}_i,\set{w}\rangle -b  \leq -1, & \text{for $y_i = -1$} \\
    \langle\set{x}_i,\set{w}\rangle -b  \geq 1, & \text{for $y_i = 1$} .
    \end{array} 
\right.
\end{equation}
To determine the margin surrounding the separating hyperplane, 
let us consider support vectors $\set{x}_{+}$ and $\set{x}_{-}$ that belong to classes $+1$ and $-1$, correspondingly. The margin can be found as a projection of a vector onto the unit normal vector of a hyperplane as
\begin{equation}
\begin{array}{r}
    \left | \left\langle (\set{x}_{+} - \set{x}_{-}), \frac{\set{w}}{\|\set{w}\|_2} \right\rangle \right |
    = \frac{ | \langle \set{x}_{+},\set{w} \rangle - \langle \set{x}_{-},\set{w} \rangle |} {\|\set{w}\|_2} \\
    =\frac{  (b+1) - (b-1)} {\|\set{w}\|_2} = \frac{2}{\|\set{w}\|_2}.
\end{array}
\end{equation}
We can determine the parameters $\set{w}$ and $b$ maximizing the margin $\frac{2}{\|\set{w}\|_2}$ between two classes by solving the following optimization problem:
\begin{flalign}
    &&\min_{\set{w},b} \quad & \frac{1}{2}\|\set{w}\|_2^2 && \label{eq:hard_margin_svm} \\
    &&\mathrm{s.t.} \quad & y_i\cdot\left[\langle\set{x}_i,\set{w}\rangle - b\right] \ge 1. \notag&&
\end{flalign}

\noindent The problem formulation in \eqref{eq:hard_margin_svm} is known as \textit{hard-margin} SVM since no training point is allowed to violate the margin constraint, which is only possible in the case of linear separability of the data. 
If the classes are linearly non-separable, we may introduce slack variables $\boldsymbol{\zeta} = \{\zeta_i\}$ to allow margin violations:
\begin{flalign}
    &&\min_{\set{w},b, \boldsymbol{\zeta}} \quad & \frac{1}{2}\|\set{w}\|_2^2 + {C}\sum_{i=1}^N \zeta_i && \label{eq:soft_margin_svm} \\
    &&\mathrm{s.t.} \quad & y_i\cdot\left[\langle\set{x}_i,\set{w}\rangle - b\right] \ge 1 - \zeta_i, \quad \zeta_i\geq 0, \notag&&
\end{flalign}
where the regularization coefficient $C$ is a hyperparameter that controls the penalty for violating the margin condition. The problem formulation in \eqref{eq:soft_margin_svm} is known as \textit{soft-margin} SVM or $C$-SVM, where the margin constraints are relaxed and the margin violation penalty is controled by the regularization coefficient $C$. 

The above optimization problem is convex and can be solved either by applying gradient descent-based methods directly or by solving the dual problem using the method of Lagrange multipliers. For the gradient descent-based optimization, it is often more convenient to reformulate the problem in \eqref{eq:soft_margin_svm} as an equivalent Tikhonov regularization problem:
\begin{flalign}
    && \min_{\set{w},b} \quad & \frac{1}{2C}\|\set{w}\|_2^2 + \sum_{i=1}^N \left[1 - y_i \cdot (\langle\set{x}_i,\set{w}\rangle - b)\right]_+, && \label{eq:csvm_tikhonov}
\end{flalign}
where $[x]_+=\max(0,x)$ is the hinge loss function.

\section{Relation between HDC and SVM}
\label{sec:hdc-svm}

In this section, we formulate the primal and dual optimization problems for binary HDC classification and demonstrate their alignment with the SVM optimization framework. We further use this relation to show that the perceptron-based HDC retraining procedure in \eqref{eq:upd_rule} is equivalent to performing gradient descent-based optimization of the formulated HDC loss function.

\subsection{Primal HDC Optimization Problem} 
\label{sec:hdc_optimization_formulation}

Let $\set{p}_-$ and $\set{p}_+$ be the prototypes of two classes with labels $-1$ and $+1$, respectively. In a binary HDC classifier, we infer that a data point $\set{x}_i$ belongs to a class $\mathcal{C}_+$ or $\mathcal{C}_-$  if it has the highest similarity with the corresponding prototype. Formally, the data point $\set{x}_i$ is classified correctly if 
\begin{equation}\label{eq:decision_rule}
    y_i\cdot\left[\mathrm{sim}(\theta(\mathbf{x}_i),\set{p}_+) - \mathrm{sim}(\theta(\mathbf{x}_i),\set{p}_-)\right] > 0, 
\end{equation}
where $y_i$ is the label of $\set{x}_i$. {For simplicity, we assume dot product--based similarity $\mathrm{sim}(\set{a},\set{b})={\langle\set{a},\set{b}\rangle}$. Our analysis is also valid for a similarity measure based on the dot product normalized by $\|\theta(\set{x}_i)\|_2$ and can be easily extended to the case of cosine similarity $\mathrm{sim}(\set{a},\set{b})=\frac{\langle\set{a},\set{b}\rangle}{\|\set{a}\|_2\|\set{b}\|_2}$.} 

{To demonstrate the connection between the binary SVM and HDC classifiers, we rewrite the condition in \eqref{eq:decision_rule} as}
\begin{equation}\label{eq:2}
y_i\cdot \left\langle\theta(\set{x}_i),{\set{p}_+} - {\set{p}_-}\right\rangle > 0.
\end{equation}
Let us consider the difference between the prototypes $\set{p}_+$ and $\set{p}_-$ in \eqref{eq:2} as a separating hyperplane ${\set{w} = \set{p}_+ - \set{p}_-}$. Consequently, the condition in \eqref{eq:2} can be reformulated as
\begin{equation}\label{eq:2_1}
y_i\cdot\langle\theta(\set{x}_i),\set{w}\rangle > 0,
\end{equation}
which translates into the decision rule $\hat y_i = \operatorname{sign} \langle\theta(\set{x}_i),\set{w}\rangle$, identical to the SVM decision rule in \eqref{eq:svm_decision_rule} with a zero bias term. 

The classifier depends only on the sign of $\langle\theta(\set{x}_i),\set{w}\rangle$. Therefore, we can rescale the parameter vector $\set{w}$ such that the closest training point corresponds to the unit margin, similarly to SVM. Following this normalization, the condition in \eqref{eq:2_1} is transformed into the equivalent inequality
\begin{equation}\label{eq:3_slack}
    y_i\cdot\langle\theta(\set{x}_i),\set{w}\rangle \ge 1,
\end{equation}
which turns into equality only for the support vectors $\theta(\set{x}_j)$ that determine the prototypes $\set{p}_+$ and $\set{p}_-$, as we demonstrate further on.

The expression in \eqref{eq:3_slack} defines the constraints that the HDC model should satisfy. Based on these constraints, we formulate an optimization problem to improve the generalization ability of the resulting HDC model. To formulate the objective function, we search for a separating hyperplane $\set{w}$ that maximizes the margin between two classes. The said margin is equal to $\frac{2}{\|\set{w}\|_2}$ after the rescaling applied in \eqref{eq:3_slack}. The objective function can be formulated by analogy with the hard-margin SVM. To address classes that cannot be linearly separated, we add slack variables $\boldsymbol{\zeta} = \{\zeta_i\}, \zeta_i \geq 0$. The optimal value of $\set{w}$ can then be found by solving the following minimization problem:
\begin{flalign}
    &&\min_{\set{w},\boldsymbol{\zeta}} \quad & \frac{1}{2}\|\set{w}\|_2^2 + {C}\sum_{i=1}^N\zeta_i && \label{eq:5} \\
    &&\mathrm{s.t.} \quad & y_i\cdot\langle\theta(\set{x}_i),\set{w}\rangle \ge 1 - \zeta_i, \quad \zeta_i \geq 0. && \nonumber
\end{flalign}

The problem in \eqref{eq:5} matches the optimization problem for the linear soft-margin SVM with a zero-bias term $b$. The key advantage of solving the formulated optimization problem over the conventional perceptron-based retraining procedures is the improved generalization capability of the HDC model. 
{According to statistical learning theory, the model generalization ability is closely linked to the model complexity, which in SVM is inversely related to the margin and can be characterized by the norm of the weight vector. This complexity is controlled by the regularization term $\frac{1}{2}\|\set{w}\|^2_2$, which promotes large margins, and by the regularization constant $C$ in \eqref{eq:5}, which sets the trade-off between the margin maximization and penalizing training errors.}
{In Section \ref{sec:mm-hdc}, we show that our method is a general case of the perceptron-based HDC classifier and, therefore, may have higher predictive performance compared to conventional HDC methods.} 

\textbf{Importance of formulated HDC optimization}. In practice, if a classification problem is not linearly separable under the selected HDC mapping $\theta(\cdot)$, then the perceptron-based HDC classifier may not converge and, in the worst case, ``explode'' such that the performance decreases over time. Alternatively, if the selected HDC mapping does induce linear separability, then the HDC classifier will converge to \textit{some} solution that is the first to satisfy the condition in \eqref{eq:decision_rule} for every data point. In contrast, the HDC optimization problem formulated in \eqref{eq:5} is \textit{guaranteed} to converge to the global optimum, which often provides better generalization due to margin regularization. 
Later in this work, we provide an iterative algorithm to solve this optimization problem and find $\set{p}_+$ and $\set{p}_-$ in a way similar to the perceptron procedure. 

\nnew{The formulated optimization problem demonstrates that the key criterion for selecting the transform $\theta(\cdot)$ is improved \textit{linear separability} of classes. The key corollary is that, for classification tasks (and probably for other pattern recognition tasks), HDC does not require a randomization mechanism per se. The projection of data into randomized, near-orthogonal hypervectors was inherited from VSAs and Kanerva's sparse distributed memories (SDMs) \cite{kanerva1992sparse} to support formal reasoning and information retrieval, respectively. This conclusion was illustrated in \cite{khaleghi2020prive}, where the prototype positions were selectively pruned to impede data retrieval, while retaining comparable classification performance.}

\nnew{We note that the goal of the randomization mechanism is to introduce redundancy that enhances the resilience of hypervectors to noise. The randomization mechanism can be applied separately, for example, via random projection as done in several previous works \cite{dutta2022hdnn,Zhao_Thomas_Brin_Yu_Rosing_2025}. In fact, the improved fault-tolerance was one of the motivations behind the use of high-dimensional, randomized data vectors in SDMs \cite[Sec. 1.1, Sec. 9]{kanerva1992sparse}. In the next section, we demonstrate that the optimal class prototypes are formed solely of those hypervectors that are support vectors for the selected transform $\theta(\cdot)$.}
\vspace{-1em}

\subsection{Dual HDC Optimization Problem}

Here, we present the dual formulation of the HDC optimization problem in \eqref{eq:5}. We use the standard SVM derivation with $b=0$. Solving the dual HDC optimization problem reveals that the optimized HDC prototypes $\set{p}_+$ and $\set{p}_-$ may be represented as a linear combination of the support vectors from the corresponding classes. We solve the dual HDC optimization problem using the method of Lagrange multipliers. This method involves maximizing the Lagrangian function formulated based on the constrained optimization in \eqref{eq:5} w.r.t. Lagrange multipliers $\boldsymbol{\lambda}=\{\lambda_i\}$ and $\boldsymbol{\eta}=\{\eta_i\}$.
The corresponding saddle-point optimization problem can be formulated as follows:
\begin{flalign}
    \max_{\boldsymbol{\lambda}, \boldsymbol{\eta}} \left[ \vphantom{\sum_{i=1}^N} \min_{\set{w},\boldsymbol{\zeta}} \right. & \left. \frac{1}{2}\|\set{w}\|_2^2 - \sum_{i=1}^N \lambda_i \left(y_i \cdot\langle\theta(\set{x}_i),\set{w}\rangle - 1\right)\right. && \notag \\
    & \left. - \sum_{i=1}^N \zeta_i\left(\lambda_i + \eta_i - C\right) \right], & \label{eq:lagrange_eq}\\
   \mathrm{s.t.}\qquad  
   & \lambda_i \ge 0, \quad \eta_i \ge 0, \quad \zeta_i \ge 0, && \\
    &   y_i \cdot \langle\theta(\set{x}_i),\set{w}\rangle -1 + \zeta_i \geq 0  ,&&\\
   &  \lambda_i \left( y_i \cdot \langle\theta(\set{x}_i),\set{w}\rangle -1 + \zeta_i\right) = 0  ,&&\\
   &  \eta_i \zeta_i = 0 \text{ for all } i=1,...,N .&&
   \label{eq:kkt3}
\end{flalign}
We denote the Lagrangian as $\mathcal L(\set{w}, \boldsymbol{\zeta}, \boldsymbol{\lambda}, \boldsymbol{\eta})$. For the sought saddle point, the following conditions hold: 
\begin{flalign}
\hspace{-1em} \frac{\partial \mathcal L ( \set{w},  \boldsymbol{\zeta}, \boldsymbol{\lambda}, \boldsymbol{\eta})}{\partial \set{w}} = 0,  \frac{\partial\mathcal{L( \set{w},  \boldsymbol{\zeta}, \boldsymbol{\lambda}, \boldsymbol{\eta})}}{\partial\zeta_i} = 0, i=1,...,N. & \label{eq:deriv}
\end{flalign}
The zero derivatives in \eqref{eq:deriv} yield two additional conditions:
\begin{equation}
\hspace{-10em}
\left\{
   \begin{array}{l}
   \set{w} = \sum_{i=1}^N \lambda_i y_i \theta(\set{x}_i),  
   \\
   \eta_i + \lambda_i = C, \quad i=1,...,N.  \label{eq:wc}
\end{array} \right.
\end{equation}

The separating hyperplane is defined above as a difference between the class prototypes $\set{p}_+$ and $\set{p}_-$, i.e., ${\set{w}=\set{p}_+ - \set{p}_-}$. Hence, the form of the first equation in \eqref{eq:wc} allows us to express each prototype as a sum of the support vectors weighed with the corresponding non-zero Lagrange multipliers:
\begin{equation}\label{eq:w_diff}
\!\set{w}\! = \! \!\sum_{i=1}^N\!\lambda_iy_i\theta(\set{x}_i) \! =  \!\!\!\!\sum_{\set{x}_i\in\mathcal{C}_+}\! \!\lambda_i\theta(\set{x}_i) - \!\sum_{\set{x}_j\in\mathcal{C}_-}\!\! \lambda_j\theta(\set{x}_j) \!= \!\set{p}_+\!-\set{p}_-,
\end{equation}
where sets $\mathcal{C}_+$ and $\mathcal{C}_-$ contain the data points corresponding to the classes $+1$ and $-1$, respectively.
{Therefore, the conventional prototype construction procedure in \eqref{eq:1} essentially initializes Lagrange coefficients, which are then iteratively refined later on during the HDC retraining procedure. Interestingly, in very high-dimensional regimes typical to HDC, large fraction of data points may become support vectors with high probability \cite{hsu2021proliferation}. This can serve as a formal rationale for bundling all data points into class prototypes before their refinement.}

Furthermore, substituting the equations in \eqref{eq:wc} into the Lagrange function, we eliminate all variables $\eta_i$ and $\zeta_i$ from the optimization problem. As a result, we obtain the following dual problem formulation:
\begin{flalign}
    &&\max_{\boldsymbol{\lambda}} \quad & \sum_{i=1}^N\lambda_i - \frac{1}{2} \sum_{i=1}^N\sum_{j=1}^N \lambda_i \lambda_j y_i y_j \langle\theta(\set{x}_i),\theta(\set{x}_j)\rangle && \label{eq:lagrange_dual}\\
    &&\mathrm{s.t.} \quad & 0 \le \lambda_i \le C, \quad i=1,...,N,\label{eq:lambda_cond} &&
\end{flalign}

\noindent
where the constraint in \eqref{eq:lambda_cond} follows from the second expression in \eqref{eq:wc}. Solving the optimization problem above provides the optimal Lagrange coefficients $\boldsymbol{\lambda}$ determining the weights of the support points to be bundled into the class prototypes $\set{p}_+$ and $\set{p}_-$  as demonstrated in \eqref{eq:w_diff}. 

\subsection{Maximum-Margin HDC Classifier}
\label{sec:mm-hdc}

Here, we devise a procedure to train the proposed maximum-margin HDC (MM-HDC) classifier. With the established relation in mind, we reformulate the primal optimization problem in \eqref{eq:csvm_tikhonov} as 
\begin{flalign}
  \!\! \min_{\set{p}_+,\set{p}_-} \! \frac{1}{2C}\|\set{p}_+ \! - \!\set{p}_-\|_2^2 \! + \! \sum_{i=1}^N \left[1 \!- \!y_i \! \cdot \! \langle\theta(\set{x}_i),\set{p}_+\! \!- \!\set{p}_- \!\rangle\right]_+\!. \!\! \label{eq:hdc_tikhonov}
\end{flalign}
Further, we adapt the approach from \cite{ergun2023federated,tian2025federated} to design an iterative retraining procedure based on the batched gradient descent algorithm. Using this algorithm, we iteratively optimize the parameters $\set{p}_+$, and $\set{p}_-$ by applying the following update rule: 
\begin{flalign}
    \set{p}_+^{(t+1)}= \set{p}_+^{(t)} - \alpha\cdot\frac{\partial \mathcal{F}}{\partial \set{p}_+},\ \set{p}_-^{(t+1)}= \set{p}_-^{(t)} - \alpha\cdot\frac{\partial \mathcal{F}}{\partial \set{p}_-}, \label{eq:gd_hdc_positive}
\end{flalign}
where $\mathcal{F}(\cdot)$ is the objective function in \eqref{eq:hdc_tikhonov} and $t$ is the optimization step index. 

We group class $+1$ and $-1$ points that are either misclassified or classified correctly but located inside the margin into sets $\mathcal{A}_+$ and $\mathcal{A}_-$, correspondingly. Particularly, $x_i \in \mathcal{A}_+ \cup \mathcal{A}_-$ if and only if $1-y_i \langle \theta(\set{x}_i), w\rangle > 0$. In other words, these sets combined contain all the data points for which the condition in \eqref{eq:3_slack} does not hold. 
We calculate the derivatives of $\mathcal{F}$ as follows:
\begin{flalign}
    & \frac{\partial \mathcal{F}}{\partial \set{p}_+} = \frac{1}{C}(\set{p}_+ \!-\! \set{p}_-) \!- \!\left(\sum_{\set{x}_i\in\mathcal{A}_+}\!\!\!\theta(\set{x}_i) \!- \!\! \!\sum_{\set{x}_j\in\mathcal{A}_-}\!\!\theta(\set{x}_j)\right) \!\!, \!\label{eq:gd_positive_deriv} \\
    & \frac{\partial \mathcal{F}}{\partial \set{p}_-} = \frac{1}{C}(\set{p}_- \!-\! \set{p}_+) \!-\! \left(\sum_{\set{x}_i\in\mathcal{A}_-}\!\!\!\theta(\set{x}_i) \!- \! \!\!\sum_{\set{x}_j\in\mathcal{A}_+}\!\!\theta(\set{x}_j)\right)\!. \! \label{eq:gd_negative_deriv}
\end{flalign}
Substituting the derivatives in \eqref{eq:gd_positive_deriv} and \eqref{eq:gd_negative_deriv} into the respective equations in \eqref{eq:gd_hdc_positive}, we complete the iterative update procedure for MM-HDC retraining. 

The introduced retraining procedure can be easily extended to the case of cosine similarity. For inequality \eqref{eq:2_1} to hold, the prototypes should have equal $\ell_2$-norms. In this case, the problem in \eqref{eq:hdc_tikhonov} can be solved using projected gradient descent, where after each update, the prototypes are normalized to preserve the condition $\|\set{p}_+\|_2=\|\set{p}_-\|_2$. {This method can be extended to online learning by applying the update rule in \eqref{eq:gd_hdc_positive} to batches of streamed data. In online setting, parameters can be further tuned to provide better adaptability.}

Based on the form of derivatives in \eqref{eq:gd_hdc_positive}, we may observe that the conventional perceptron-based HDC classifier is a special case of the derived algorithm with $C\to\infty$. This implies that the key advantage of the MM-HDC classifier is in its inherent regularization support that may enhance the generalization ability of HDC. We note that the derivation above may not be limited to batched gradient descent. The update rule in \eqref{eq:gd_hdc_positive} could be modified to support more advanced gradient descent-based methods similar to the one in~\cite{kim2024advancing}. 

{Currently, the MM-HDC update procedure supports only real-valued HD transforms. 
Our algorithm, however, can be implemented in fixed-point arithmetic with a specified bitwidth similar to \cite{hernandez2021onlinehd}. Essentially, our MM-HDC method offers an ``HDC way'' of linear $C$-SVM training with batched gradient descent. With this reformulation, our work provides a mechanism to introduce margin-based learning into HDC \nnew{while preserving implementation simplicity that originally motivated HDC research.}}

\begin{figure*}[t!]
    \centering
    \captionsetup[subfloat]{labelformat=empty}
    \subfloat[\scriptsize(a) MNIST dataset]{%
        \includegraphics[width=1.98in]{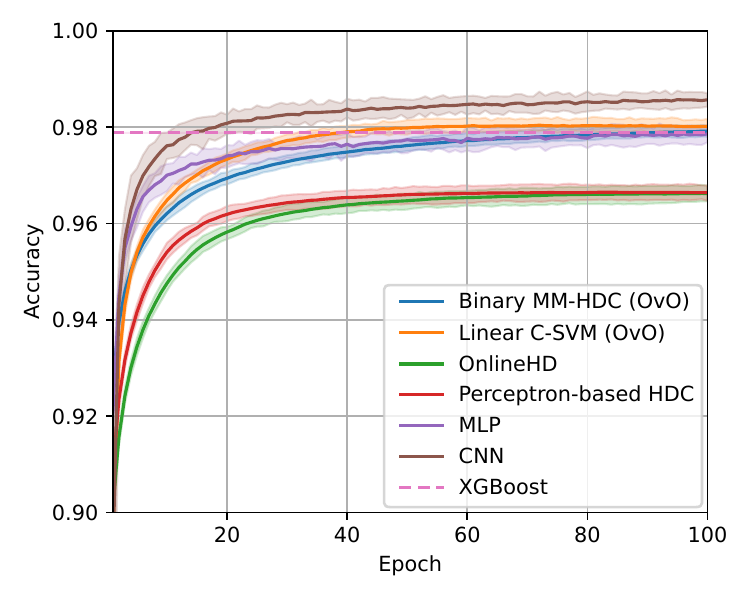}%
        \label{fig:sim_results_1}
    }
    \hfil
    \subfloat[\scriptsize(b) Fashion MNIST dataset]{%
        \includegraphics[width=1.98in]{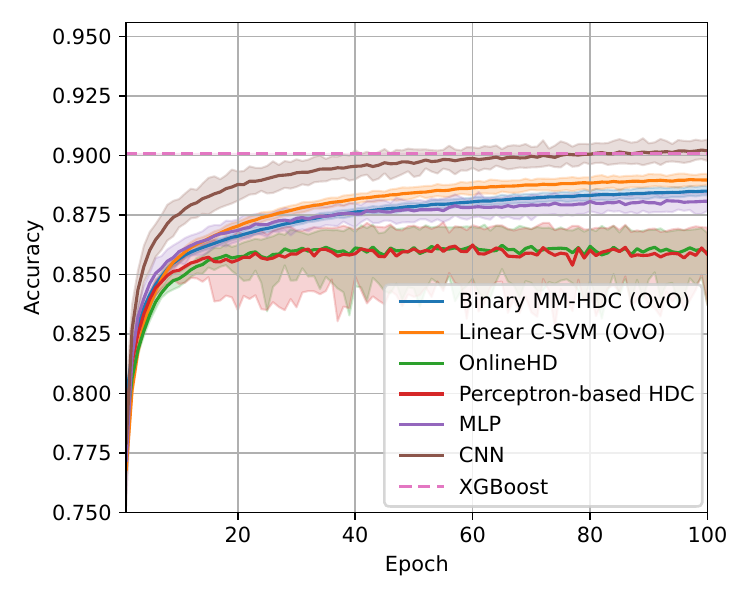}%
        \label{fig:sim_results_2}
    }
    \hfil
    \subfloat[\scriptsize(c) UCI HAR dataset]{%
        \includegraphics[width=1.98in]{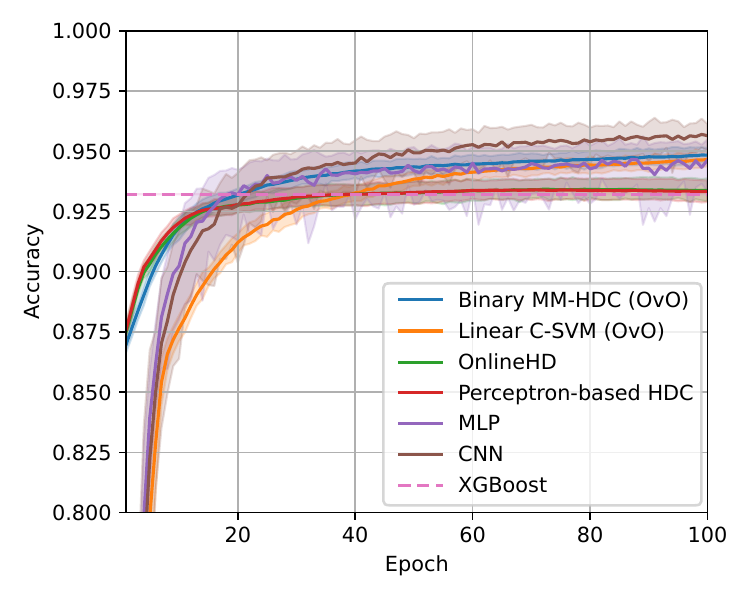}%
        \label{fig:sim_results_3}
    }
    \caption{Comparison of performance of maximum-margin HDC (blue), linear $C$-SVM (orange), OnlineHD (green), perceptron-based HDC (red), and other algorithms.}
    \label{fig:sim_results}
    \vspace{-3mm}
\end{figure*}

\subsection{Complexity Analysis}
{Here, we analyze the computational complexity of our MM-HDC method. For convenience, we provide an assessment for every key procedure of the algorithm: prototype initialization, similarity check, and prototype retraining. We express the costs in the number of elementary operations (addition, subtraction, multiplication, division) and assume that all operations have equal complexity. We also exclude the projection of the data into hypervectors from the analysis, as our method is agnostic to the selection of the HD transform.}

{\textbf{Prototype initialization.} In our framework, the bundling operation in \eqref{eq:1} is implemented as an element-wise summation. Since the prototype initialization procedure bundles all data points into the corresponding class prototypes, the computational cost of this step is $ND$.}

{\textbf{Similarity check.} In our method, the similarity check involves computing dot products between the test point $\set{x}_*$ and the positive and negative class prototypes $\set{p}_+$ and $\set{p}_-$, respectively. The computational complexity of the dot product between two $D$-dimensional hypervectors is $2D-1$. Therefore, the total inference complexity is ${2\cdot(2D-1)\sim \mathcal{O}(D)}$.}

{\textbf{Prototype retraining.} Recall that the retraining procedure of MM-HDC consists of iteratively updating the prototypes $\set{p}_+$ and $\set{p}_-$ according to the corresponding update rules in \eqref{eq:gd_positive_deriv} and \eqref{eq:gd_negative_deriv}. Let us assume that the retraining procedure is performed over the batch of training points of size $B$, each belonging to either set $\mathcal{A}_+$ or set $\mathcal{A}_-$. One can observe that the derivative terms in \eqref{eq:gd_positive_deriv} and \eqref{eq:gd_negative_deriv} differ only in sign and, therefore, can be precomputed. Given this observation, the computational complexity of updating both prototypes is ${(B + 7)D + 1\sim\mathcal{O}(BD)}$.}

\begin{table}[t!]
    \centering
    {\caption{Parameters of considered HDC and SVM methods.}
        \begin{tabular}{|l|c|c|c|}
        \hline
        \multirow{2}{*}{Hyperparameter} & \multicolumn{3}{c|}{Dataset} \\
        \cline{2-4}
        &  MNIST & FASHION & UCI HAR \\
        \hline
        \hline
        \multicolumn{4}{|c|}{\textbf{All methods}} \\
        \hline
        HD representation size, $D$ & \multicolumn{3}{c|}{$5000$} \\
        \hline
        Batch size & \multicolumn{3}{c|}{$1000$} \\ 
        \hline
        \multicolumn{4}{|c|}{\textbf{Maximum-margin HDC}} \\
        \hline
        Learning rate, $\alpha$ & \multicolumn{3}{c|}{$1\cdot 10^{-5}$} \\
        \hline
        Regularization constant, $C$ & \multicolumn{3}{c|}{$500$}\\ 
        \hline
        \multicolumn{4}{|c|}{\textbf{Linear $C$-SVM}} \\
        \hline
        Learning rate, $\alpha$ & \multicolumn{3}{c|}{$1\cdot10^{-4}$} \\
        \hline
        Regularization constant, $C$ & \multicolumn{3}{c|}{$500$}\\ 
        \hline
        \multicolumn{4}{|c|}{\textbf{Perceptron-based HDC and OnlineHD}} \\
        \hline
        Learning rate, $\alpha$ & \multicolumn{3}{c|}{$1\cdot 10^{-5}$} \\
        \hline
    \end{tabular}
    \label{tab:parameters}
    }
    \vspace{-1em}
\end{table}

\section{Numerical Results}
In this section, we fix the mapping $\theta(\set{x})$, which makes MM-HDC equivalent to a linear soft-margin SVM formulation with zero bias. The main objective of the experiments below is to study the benefits of explicit margin-based optimization compared to conventional HDC update rules. 
\subsection{Experimental Setup}

\begin{table*}[ht]
\centering
\caption{Classification performance comparison of MM-HDC with the baselines in terms of mean accuracy scores at the last epoch.\\}
{
\begin{tabular}{|l|c|c|c|c|c|}
\hline
\multirow{2}{*}{\textbf{Classifier}} & \multirow{2}{*}{\textbf{Enc. type}} & \multirow{2}{*}{\textbf{Format}} & \multicolumn{3}{c|}{\textbf{Accuracy}} \\ \cline{4-6} 
                                  & & &  \textbf{MNIST} & \textbf{FASHION} & \textbf{UCI HAR} \\ \hline\hline
    \multicolumn{6}{|c|}{\textbf{HDC methods}} \\ \hline               
    Binary MM-HDC (OvO) & NL & FP32 & 0.979 & 0.885 & 0.948 \\ \hline
    OnlineHD & NL & FP32 & 0.966 & 0.859 & 0.934 \\ \hline
    Perceptron-based HDC & NL & FP32 & 0.966 & 0.858 & 0.933 \\ \hline
    \nnew{FedHDC \cite{ergun2023federated, tian2025federated}} & \nnew{NL} & \nnew{BP/FP32} & \nnew{0.968} & \nnew{-} & \nnew{0.945} \\ \hline
    TrainableHD \cite{kim2024advancing} & NL & \nnew{INT8/FP32} & 0.980 & - & 0.955 \\ \hline
    DependableHD \cite{liang2023dependablehd} & RP & \nnew{INT16} & 0.966 & - & 0.969 \\ \hline
    LeHDC \cite{duan2022lehdc} & BNN & \nnew{BP/FP32} & 0.947 & 0.871 & 0.952 \\ \hline
    Laplace-HDC \cite{pourmand2025laplace} & CB & \nnew{BP/FP32} & 0.964 & 0.887 & - \\ \hline
    RFF-HDC \cite{yu2022understanding} & CB & \nnew{BP/FP32} & 0.966 & 0.874 & 0.966 \\ \hline
    \multicolumn{6}{|c|}{\textbf{Deep learning and classical ML methods}} \\ \hline
    Linear $C$-SVM & - & - & 0.980 & 0.890 & 0.947 \\ \hline
    MLP & - & - & 0.979 & 0.881 & 0.946 \\ \hline
    CNN & - & - & 0.986 & 0.902 & 0.956 \\ \hline
    XGBoost & - & - & 0.979 & 0.901 & 0.932 \\ \hline
\end{tabular}
}
\label{tab:performance}
\end{table*}

We evaluate the performance of our MM-HDC algorithm on several community-adopted multi-class classification datasets, including MNIST \cite{lecun1998mnist}, Fashion MNIST (FASHION) \cite{xiao2017fashion}, and {UCI HAR} \cite{anguita2013public}.
For each of these datasets, we apply the HD transform proposed in \cite{hernandez2021onlinehd}, i.e.,
\begin{equation}
   \theta(\set{x}) = \cos(\set{x}\set{W}+\boldsymbol{\varphi})\cdot\sin(\set{x}\set{W}),
   \label{eq:onlinehd_transform}
\end{equation}
where $\set{W}\sim\mathcal{N}(\set{0},\set{I})$ is a $d\times D$ random projection matrix and $\boldsymbol{\varphi}\sim\mathrm{Uni}[0,2\pi]$ is a random $D$-dimensional phase shift vector. {This type of encoding is often referred to as non-linear encoding \cite{kim2024advancing}, which can be interpreted as a fully connected layer of a neural network with a non-linear function applied to its output. The parameters $\set{W}$ and $\boldsymbol{\phi}$ can be sampled using a pseudo-random number generator with some predefined seed, as done in many conventional HDC frameworks. \nnew{The OnlineHD mapping has been successfully adopted in previous HDC frameworks for image retrieval \cite{yun2024neurohash}, image classification \cite{ ni2022neurally}, and time series forecasting \cite{moreno2024kalmanhd}, which confirms its universality for a wide range of applications.}}

To control feature space, we also normalize the individual data points to a unit $\ell_2$-norm prior to applying the HD transform in \eqref{eq:onlinehd_transform}. Based on our previous discussion, we note that the HD transform $\theta(\set{x})$ acts as a non-trainable feature transform and can be replaced by a pretrained neural network-based feature extractor, as demonstrated in~\cite{dutta2022hdnn}.
{Ultimately, the HD transform $\theta(\cdot)$ can be \textit{any} function that transforms the original data $\set{x}$ into a suitable feature space, not necessarily high-dimensional. A key insight of our work is that the feature transform $\theta(\cdot)$ does not require high dimensionality, or randomness for HDC to function efficiently. Randomized feature expansions provide a set of advantageous properties, such as near-orthogonality and resilience to noise \cite{hernandez2021onlinehd}.} 
\nnew{We note that randomization is not strictly required. The key requirement is that the HD transform $\theta(\cdot)$ maps the data into a space where classes are linearly separable. }

\subsection{Comparison with Baseline Algorithms}

We compare our MM-HDC algorithm to several baseline models: linear $C$-SVM \cite{cortes1995support}, perceptron-based HDC, OnlineHD \cite{yu2022understanding}, \nnew{FedHDC \cite{ergun2023federated, tian2025federated},} TrainableHD \cite{kim2024advancing}, DependableHD \cite{liang2023dependablehd}, LeHDC \cite{duan2022lehdc}, Laplace-HDC \cite{pourmand2025laplace}, and RFF-HDC \cite{yu2022understanding}. 
Here, perceptron-based HDC corresponds to the basic HDC method with the conventional perceptron-based retraining procedure as in \eqref{eq:upd_rule}. 
\nnew{The FedHDC method is based on the Fisher's linear discriminant and, therefore, is conceptually closest to our framework. We also note that the reported performance of FedHDC is for the federated implementation and, therefore, may be higher for the centralized case.}

\nnew{For linear $C$-SVM, perceptron-based HDC, and OnlineHD, we report the performance results obtained from own implementations. 
For other baseline methods, we reproduce the classification performance results reported in the original publications. As the hyperparameter values, such as the hypervector size $D$, can differ between the implementations, we include the highest reported performance figures. We note that all baseline methods from the referenced works employ bipolar hypervectors of size $D=10000$, except for DependableHD, which employs integer-valued hypervectors of size $D=3000$.}

{The baseline HDC methods rely on different encoding algorithms and hypervector data formats. There are three main types of employed HD transforms: non-linear (NL), random projection (RP)-based, and codebook (CB)-based. The former two are based on the matrix-to-vector multiplication between the data vector and the randomized short-and-fat matrix. The only difference is that in the RP-based transform, the matrix is sampled from a Bernoulli distribution, while the non-linear transform utilizes a Gaussian matrix and applies non-linearity to the matrix-to-vector product. The codebook-based transforms rely on conventional symbolic HDC mappings based on the algebra over hypervectors. The only exception is the LeHDC method \cite{duan2022lehdc} that employs a binary neural network (BNN) for feature extraction.}

\begin{figure*}[ht]
    \centering
    \captionsetup[subfloat]{labelformat=empty}
    \subfloat[\scriptsize(a) Size of hypervector $D=500$]{%
        \includegraphics[width=1.98in]{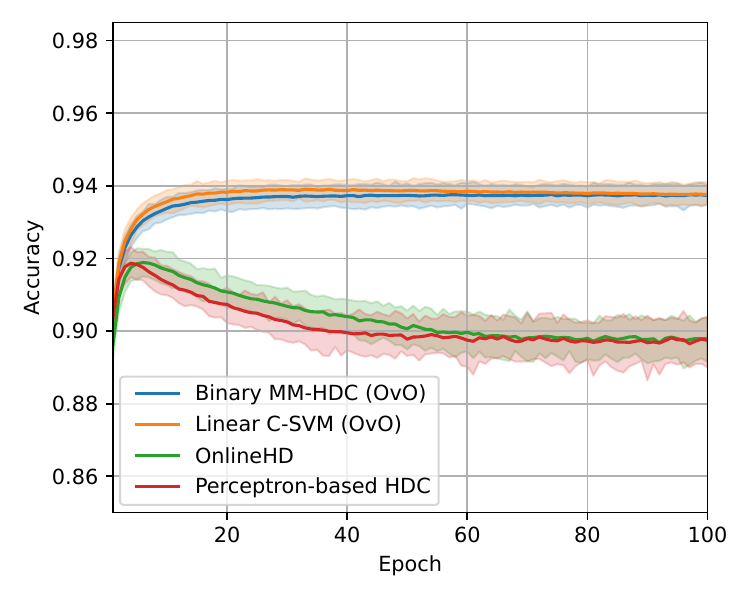}%
        \label{fig:pr_results_1}
    }
    \hfil
    \subfloat[\scriptsize(b) Size of hypervector $D=1000$]{%
        \includegraphics[width=1.98in]{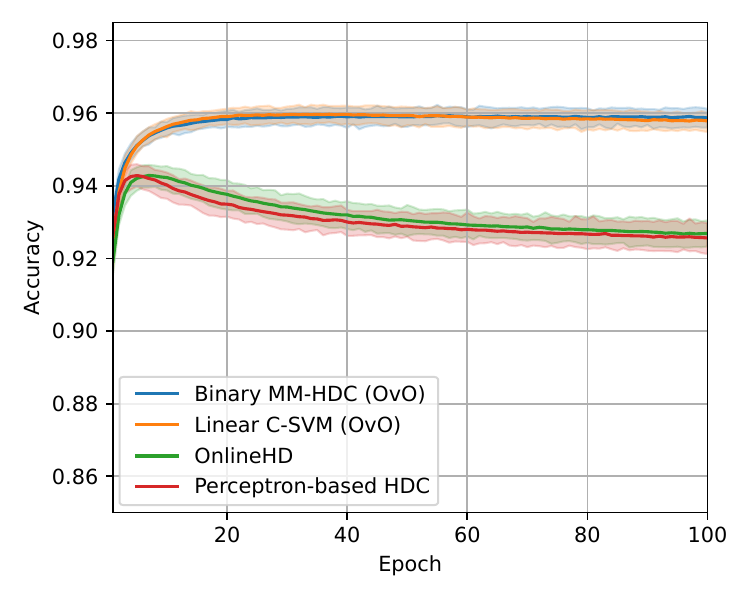}%
        \label{fig:pr_results_2}
    }
    \hfil
    \subfloat[\scriptsize(c) Size of hypervector $D=2500$]{%
        \includegraphics[width=1.98in]{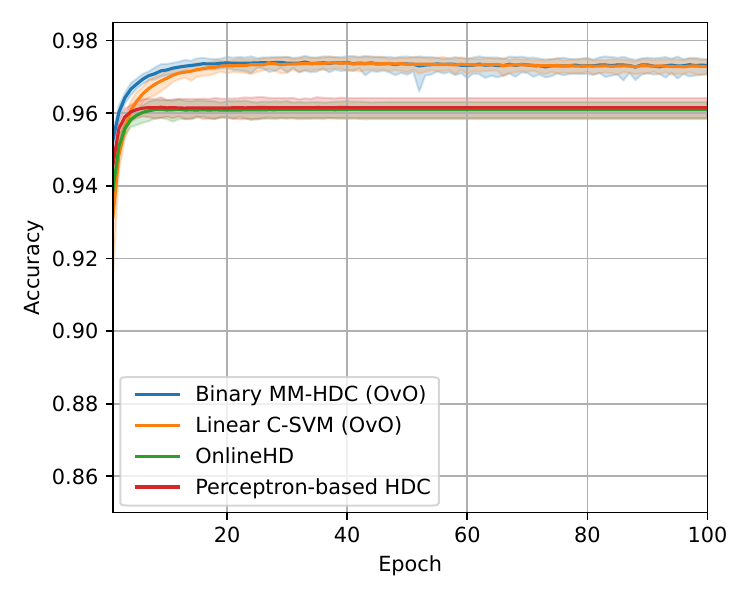}%
        \label{fig:pr_results_3}
    }
    \caption{Performance comparison of maximum-margin HDC (blue), linear $C$-SVM (orange), OnlineHD (green), and perceptron-based HDC (red) for MNIST dataset and different sizes of hypervectors $D$.}
    \label{fig:pr_results}
    \vspace{-3mm}
\end{figure*}

In our implementation, the linear $C$-SVM, perceptron-based HDC, and OnlineHD use a variation of the non-linear transform, namely, the OnlineHD transform in \eqref{eq:onlinehd_transform}, with the discussed preprocessing steps. Most of the baseline methods rely on bipolar (BP) or integer-valued (INT8/INT16) hypervectors for better resource efficiency, while we employ real-valued (FP32) hypervector representations. We also note that the performance gap between using real-valued and bipolar hypervectors can be marginal when the BP hypervectors are sufficiently large \cite{thomas2021theoretical}. 

\nnew{We note that all baselines, except for DependableHD, rely on floating-point operations during the training or inference stage. Specifically, FedHDC constructs bipolar data hypervectors, but the prototype is real-valued, and its refinement is done in the floating-point arithmetic \cite[Sec. 4]{ergun2023federated}.
The TrainableHD method adopts quantization-aware training by refining the prototypes in the FP32 format and subsequently quantizing them at the inference stage into the INT8 format \cite[Sec. 6.2]{kim2024advancing}. 
The LeHDC \cite[Sec. 4]{duan2022lehdc} and RFF-HDC \cite[Sec. 7]{yu2022understanding} methods optimize the bipolar prototypes using SGD with straight-through estimation, which involves computation of real-valued gradients. The Laplace-HDC framework similarly relies on the real-valued SGD-based training of the prototypes, the values of which are clamped to sign either every or only the last iteration, depending on the selected regime \cite[Sec. 2.1]{pourmand2025laplace}.}

Moreover, we discover that the perceptron-based HDC and OnlineHD methods may converge significantly faster when their prototypes are normalized after each retraining iteration. Therefore, we additionally implement this normalization practice for these methods in our experiments. The parameters of MM-HDC, perceptron-based HDC, OnlineHD, and linear $C$-SVM methods are summarized in Table \ref{tab:parameters}.

Since SVM and MM-HDC methods are inherently designed for binary classification, we adopt one-versus-one (OvO) classification to test them on standard multi-class datasets. For the classification problem with $K$ classes, the OvO classifier trains~$\frac{K(K-1)}{2}$ models for each pair of classes. Alternatively, one may adopt a one-versus-rest classification with only $K$ classifiers, which is more similar to the HDC framework. However, we observe that one-versus-one classification performs significantly better on the selected datasets. Consequently, we adopt the OvO approach despite its higher computational~complexity. 

{In addition, we compare our MM-HDC method with other deep learning and classical ML methods: MLP (multi-layer perceptron), CNN (convolutional neural network), and XGBoost \cite{chen2016xgboost}. For MLP, we use a three-layered configuration $(512,256,126)$ with ReLU activations. For CNN, we employ two $(32, 64)$ convolutional layers with $2\times2$ max-pooling and the last $128$-dimensional fully-connected layer. Both networks are trained with $0.2$ dropout rate, softmax outputs, cross-entropy loss, and Adam optimizer \cite{kingma2015adam} with $1\cdot 10^{-3}$ learning~rate.}

We present a comparison of the considered methods in terms of classification accuracy in Fig.~\ref{fig:sim_results} and Table~\ref{tab:performance}. For each method, we provide the mean accuracy averaged over 50 experiments, together with $(5,95)$-percentile. For SVM training, we employ the Adam optimizer since it demonstrates significantly more stable and faster convergence compared to batched gradient descent-based optimization. The results demonstrate that even with these enhancements, our MM-HDC method achieves smoother convergence with a notable accuracy improvement compared to the baseline HDC methods. 

\nnew{Slower convergence of MM-HDC compared to linear $C$-SVM can be attributed to the difference in the optimization backbones. The retraining procedure of MM-HDC is based on SGD, which demonstrates generally slower convergence compared to the Adam optimizer employed for the linear $C$-SVM training. We note that the SGD-based retraining procedure can be readily extended to the Adam-based optimization as demonstrated in \cite{kim2024advancing}. However, both methods have equivalent loss functions (excluding the bias term) and demonstrate almost identical classification performance for smaller hypervector sizes $D$ as demonstrated in Section IV.C~below.}

{In our experiments, we also observe an interesting phenomenon: for the applied OnlineHD mapping in \eqref{eq:onlinehd_transform} and the selected hypervector size $D=5000$, lowering the impact of regularization (equivalently, increasing the regularization parameter so that the objective function is less sensitive to the changes in $\|\set{p}_i\|_2$) results in higher classification accuracy. 
}

\subsection{Impact of Hypervector Size}

{Despite high resource efficiency, large sizes of hypervectors can still be difficult to store for memory-constrained architectures \cite{kang2025memhd}. Therefore, we vary the dimensionality~$D$ of hypervectors and explore its effects to lower the memory usage and, consequently, compute requirements. In Fig. \ref{fig:pr_results}, we compare the classification accuracy of the considered HDC algorithms on the MNIST dataset for varied sizes~$D$ of hypervectors. For every value of~$D$, we apply a higher learning rate of $\alpha=1\cdot10^{-4}$, while the rest of the hyperparameters for the MNIST dataset are reported in Table~\ref{tab:parameters}. \nnew{We observe that a higher learning rate results in unstable training convergence for the base hypervector size of $D=5000$. Such an instability can be potentially treated with the dropout-inspired enhancement of the retraining procedure proposed in our previous work~in~\cite{zeulin2023resource}.}}

{The results reveal an interesting trend: if the hypervector size is relatively small, conventional perceptron-based and OnlineHD methods overfit, as demonstrated by a slow accuracy decay for $D\!=\!\{500,1000\}$. In contrast, our MM-HDC algorithm and its SVM counterpart maintain a stable performance over the training process. We explain this behavior by the capability of MM-HDC/SVM to attain a larger margin, which stabilized training for low values of $D$.
}

\vspace{3mm}
\section{Implications and Open Research Directions}
Sections II and III demonstrate that HDC with dot-product similarity is equivalent to a linear soft-margin SVM classifier with zero bias. This section describes practical implications of this result by elaborating on (i) relation between $\theta(\set{x})$ and kernel methods, (ii) extensions to multi-class formulations, and (iii) alternative loss functions.

\subsection{Relation to Kernel Methods}
In classical SVMs, 
a well-established practice for handling data that are not linear separable is applying the so-called \text{kernel trick}, where the dot-products $\langle \set{x}_i,\set{x}_j\rangle$ in the finite feature space are replaced by a positive-definite kernel function $k(\set{x}_i, \set{x}_j)$ that implicitly defines the feature map. For continuous shift-invariant kernels, Bochner's theorem \cite[Sec. 4.2.1]{rasmussen2006gaussian} provides a characterization that enables randomized finite-dimensional approximations. 
A practical implication of this theorem is that such kernel functions can be approximated by dot-products in the finite feature space. 
In other words, $k(\set{x}_i,\set{x}_j) \approx \langle\phi(\set{x}_i), \phi(\set{x}_j)\rangle$, where $\phi(\cdot):\mathbb{R}^d\to\mathbb{R}^D$ is a randomized explicit feature mapping corresponding to the kernel function. A well-known example of such a mapping is the random Fourier feature mapping
    $\phi(\set{x}) = \left[\cos\left(\set{x}\set{W}\right) \sin\left(\set{x}\set{W}\right)\right]$ 
approximating the radial basis function (RBF) kernel \cite{rahimi2007random}, where $\set{W}\sim\mathcal{N}(\mathbf{0},\sigma^2\mathbf{I})$ is a $d\times D$ random projection matrix and $\sigma$ is an RBF kernel parameter. 

\nnew{Following the result in \cite[Sec. 5.2.1]{thomas2021theoretical},} our discussion highlights a formal relation between HDC and kernel methods: selecting an HDC mapping $\theta(\cdot)$ as a random feature mapping $\phi(\cdot)$ turns HDC into a kernel method with ${k(\set{x}_i,\set{x}_j)\approx\langle\theta(\set{x}_i),\theta(\set{x}_j)\rangle}$. We also note that some SVM implementations may benefit from using the cosine kernel~\cite{ben2010user}, which is conventionally applied for measuring similarity in HDC algorithms. This observation further reinforces a strong connection between HDC, SVM, and kernel methods.

\subsection{Multi-Class Classification}
In contrast to conventional HDC classifiers, the presented MM-HDC approach has limited native support for multi-class classification. Similar to the original SVM formulation, binary MM-HDC can be combined with the one-vs-one or one-vs-rest classification, which may increase training and inference complexity compared to conventional multi-class HDC methods. 

{A direct multi-class MM-HDC formulation could be obtained by applying similar derivation steps to the multi-class SVM method in~\cite{weston1998multi}, here referred to as Weston-Watkins (WW)-SVM. Particularly, the decision rule of WW-SVM coincides with the HDC decision rule in \eqref{eq:hdc_class_rule} for the case of dot-product similarity. Consequently, it could be possible to extended the derivations in Section \ref{sec:hdc-svm} to a multi-class MM-HDC formulation based on WW-SVM. Such a reformulation eliminates the need for additional ensemble methods to tackle multi-class problems. This is an important extension of binary MM-HDC, the implementation of which remains open for future research.}

\subsection{Loss Function Selection} 
The original binary SVM optimization problem in \eqref{eq:csvm_tikhonov} employs the hinge loss function with $\ell_2$-regularization. The hinge loss can be replaced with other functions, such as the squared hinge loss, which may modify the behavior of the SVM. In fact, substituting the hinge loss with the squared hinge loss in the MM-HDC objective in \eqref{eq:hdc_tikhonov} reveals an interesting relation of our method to the OnlineHD retraining procedure proposed in \cite{hernandez2021onlinehd}. The MM-HDC objective with the squared hinge loss can be rewritten~as:
\begin{flalign}
   \min_{\set{p}_+,\set{p}_-} ~ & \frac{1}{2C}\|\set{p}_+ \! - \!\set{p}_-\|_2^2 \! + \! \sum_{i=1}^N \left[1 \!- \!y_i \! \cdot \! \langle\theta(\set{x}_i),\set{p}_+ \!- \!\set{p}_- \!\rangle\right]_+^2\!.
   \label{eq:hdc_tikhonov_sq}
\end{flalign}
The partial derivatives of the objective in \eqref{eq:hdc_tikhonov_sq} w.r.t. the prototypes $\set{p}_+$ and $\set{p}_-$ have the following form:
\begin{flalign}
    & \frac{\partial \mathcal{F}}{\partial \set{p}_+} = \frac{1}{C}(\set{p}_+ \!-\! \set{p}_-) \!- \!\left(\sum_{\set{x}_i\in\mathcal{A}_+}\!\!\!\Delta_i\theta(\set{x}_i) \!- \!\! \!\sum_{\set{x}_j\in\mathcal{A}_-}\!\!\Delta_j\theta(\set{x}_j)\right) \!\!, \!\notag \\
    & \frac{\partial \mathcal{F}}{\partial \set{p}_-} = \frac{1}{C}(\set{p}_- \!-\! \set{p}_+) \!-\!\left(\sum_{\set{x}_i\in\mathcal{A}_-}\!\!\!\Delta_i\theta(\set{x}_i) \!- \! \!\!\sum_{\set{x}_j\in\mathcal{A}_+}\!\!\Delta_j\theta(\set{x}_j)\right)\!, \! \notag
\end{flalign}
where $\Delta_i = 2\cdot\left[1 \!- \!y_i \! \cdot \! \langle\theta(\set{x}_i),\set{p}_+ \!- \!\set{p}_- \!\rangle\right]_+$ is the weight coefficient. \nnew{Without margin maximization, the weight coefficient becomes ${\Delta_i=2\cdot\langle\theta(\set{x}_i),\set{p}_+-\set{p}_-\rangle}$, since $\Delta_i$ are always positive for data points $\set{x}_i$ from $\mathcal{A}_+$ and~$\mathcal{A}_-$.} Substituting the above derivatives into \eqref{eq:gd_hdc_positive} and selecting ${C\to\infty}$ results in an update rule closely resembling the OnlineHD method from \cite{hernandez2021onlinehd} \nnew{for the binary classification case}. The only difference between the methods is in the choice of the function $\Delta_i$. While OnlineHD uses a cosine distance based on heuristics, we select $\Delta_i$ as a hinge loss based on formal derivations. This suggests that adopting alternative loss functions may result in discovering multiple variations of MM-HDC tailored to different tasks.

\subsection{Neural Network-Based Feature Extraction}

{Our MM-HDC framework is compatible with the end-to-end training of the HDC classifier and neural network-based feature extractor. The core idea of such end-to-end training is to jointly optimize the encoder parameters and the HDC class prototypes using standard backpropagation. 
Particularly, let $\boldsymbol{\Theta}$ be the parameters of some neural network-based feature extractor $\theta(\set{x};\boldsymbol{\Theta})$, which maps raw input $\set{x}$ into the HD space. Gradient-based optimization can be applied to both the encoder and the HDC prototypes:
\begin{equation*}
    \set{p}^{(t+1)} = \set{p}^{(t)} - \alpha\cdot\frac{\partial\mathcal{F}}{\partial\set{p}}\qquad\text{and}\qquad \boldsymbol{\Theta}^{(t+1)} = \boldsymbol{\Theta}^{(t)} - \alpha\cdot\frac{\partial\mathcal{F}}{\partial\boldsymbol{\Theta}},
\end{equation*}
where we omit the signs of the prototypes $\set{p}_+$ and $\set{p}_-$ for brevity.
Since the HDC algorithm is a classification head for the feature extractor parameterized by $\boldsymbol{\Theta}$, its partial derivative can be represented using the chain rule:
\begin{equation*}
    \frac{\partial\mathcal{F}}{\partial\boldsymbol{\Theta}} = \frac{\partial\mathcal{F}}{\partial\set{p}_+}\cdot\frac{\partial\set{p}_+}{\partial\boldsymbol{\Theta}} + \frac{\partial\mathcal{F}}{\partial\set{p}_-}\cdot\frac{\partial\set{p}_-}{\partial\boldsymbol{\Theta}},
\end{equation*}
where the partial derivatives for the positive and negative class prototypes are derived in \eqref{eq:gd_positive_deriv} and \eqref{eq:gd_negative_deriv}, correspondingly.
This allows us to perform full end-to-end training through automatic differentiation frameworks as was previously demonstrated in \cite{duan2022lehdc, dutta2022hdnn}. If real-valued automatic differentiation cannot be implemented due to hardware constraints, it may be possible to apply binary neural networks \cite{duan2022lehdc} or fully integer neural networks that implement backpropagation without relying on floating-point arithmetic \cite{wu2018training}.}

\section{Conclusion}

In this article, we established a formal relation between binary HDC and linear soft-margin SVM classifiers. We demonstrated that a binary HDC classifier can be reformulated as a linear $C$-SVM classifier with identical decision rules. Building on this result, we formulated a convex optimization problem and proposed an iterative algorithm to train a maximum-margin HDC classifier. 
The proposed MM-HDC algorithm matches or outperforms the baseline HDC classifiers on several benchmark datasets. 

The presented analytical perspective complements the growing body of ML-oriented HDC work that focuses on encoding and representation design. Our results also provide a theoretical interpretation of several previously heuristic HDC training procedures. 
More broadly, the margin-based perspective provides a foundation for the systematic design of future HDC methods, grounded in the well-established theory of SVMs.

\bibliographystyle{ieeetr}
\bibliography{references}
\balance

\end{document}